\documentclass{article} 
\usepackage{iclr2026_conference,times}

\usepackage{amsmath,amsfonts,bm}

\def\eqref#1{equation~\ref{#1}}

\def\1{\bm{1}}

\DeclareMathAlphabet{\mathsfit}{\encodingdefault}{\sfdefault}{m}{sl}
\SetMathAlphabet{\mathsfit}{bold}{\encodingdefault}{\sfdefault}{bx}{n}

\usepackage{hyperref}
\usepackage{url}
\usepackage{booktabs}   
\usepackage{multirow}
\usepackage{adjustbox}   
\usepackage{siunitx} 
\usepackage{enumitem}

\title{Do LLMs Build Spatial World Models? \\Evidence from Grid-World Maze Tasks}

\author{Weijiang Li\\
University of Notre Dame\\
\texttt{wli27@nd.edu}
\And
Yilin Zhu\\
Columbia University\\
\texttt{yz3323@columbia.edu}
\AND
Rajarshi Das\\
MQube Cognition\\
\texttt{rajarshi.das@mqube.ai}
\And
Parijat Dube\\
Columbia University
}

\iclrfinalcopy 
\begin{document}

\maketitle

\begin{abstract}
Foundation models have shown remarkable performance across diverse tasks, yet their ability to construct internal spatial world models for reasoning and planning remains unclear. We systematically evaluate the spatial understanding of large language models through maze tasks, a controlled testing context requiring multi-step planning and spatial abstraction. Across comprehensive experiments with Gemini-2.5-Flash, GPT-5-mini, Claude-Haiku-4.5, and DeepSeek-Chat, we uncover significant discrepancies in spatial reasoning that challenge assumptions about LLM planning capabilities. Using chain-of-thought prompting, Gemini achieves 80-86\% accuracy on smaller mazes ($5 \times 5$ to $7 \times 7$ grids) with adjacency list representations, but performance collapses to 16-34\% with visual grid formats, which is a 2-5x difference, suggesting representation-dependent rather than format-invariant spatial reasoning. We further probe spatial understanding through sequential proximity questions and compositional distance comparisons. Despite achieving 96-99\% semantic coverage in reasoning traces, models fail to leverage this understanding for consistent spatial computations, indicating that they treat each question independently rather than building cumulative spatial knowledge. Our findings based on the maze-solving tasks suggest that LLMs do not develop robust spatial world models, but rather exhibit representation-specific and prompting-dependent reasoning that succeeds only under narrow conditions. These results have significant implications for deploying foundation models in applications that require spatial abstraction.
\end{abstract}

\section{Introduction}
Foundation models have emerged as powerful tools for complex reasoning tasks, demonstrating capabilities that extend beyond pattern recognition to structured problem-solving. Recent work suggests these models may develop internal ``world models", an abstract and causal representation of task structure, that enables systematic reasoning and planning~\citep{nanda_othello_2023,li2024emergentworldrepresentationsexploring,lewis2024evaluatingrobustnessanalogicalreasoning,Yildirim_2024}. However, a fundamental debate remains: Do large language models (LLMs) truly build robust world models, or do they rely on superficial pattern matching~\citep{beger2026aimodelsperformhumanlike}?  
Spatial tasks~\citep{chollet2019measureintelligence, li2024emergentworldrepresentationsexploring,feifei_2025} provide a rich environment towards exploring and addressing this question. More specifically, how model architectures, input representations, and prompting strategies underpin spatial intelligence in the spectrum of models' internal representation, from bag of heuristics to robust emergent world models~\citep{beger2026aimodelsperformhumanlike}. Understanding these mechanisms is critical as foundation models are increasingly deployed for robotics navigation, autonomous systems, and spatial planning applications. 

Spatial navigation~\citep{COPPOLINO202397,martorell2025textspacemappingabstract} provides an ideal testbed for probing world model emergence. Unlike many NLP benchmarks where success may reflect memorization of training data patterns, maze solving demands structured spatial representation of topology and connectivity, multi-step planning with sequential decision-making, compositional generalization across maze sizes and configurations, and yields clear success metrics where path correctness is objectively verifiable. If foundation models develop genuine spatial world models, we expect robust generalization to unseen configurations, consistent performance across equivalent input representations, and reasoning traces reflecting accurate spatial understanding. Conversely, brittleness to representation changes or failure to generalize would suggest surface-level pattern matching rather than abstract spatial reasoning.

This work addresses three main questions about spatial reasoning and planning in foundation models. First, regarding representation dependency, we are interested in how the input format, spatial maze represented as visual grids versus tokenized adjacency lists, affects spatial reasoning performance, and whether format sensitivity results in representation-specific heuristics rather than format-invariant spatial abstraction. Second, for few-shot learning and prompt context, we investigate whether in-context examples enhance spatial planning by instantiating task-specific world models or introduce interference through increased context length, but with a context window length as a tradeoff. Third, on reasoning transparency, we explore whether models that articulate explicit spatial reasoning demonstrate superior planning performance, and how reasoning and chain-of-thought prompting compare for spatial tasks.

\begin{figure}[h!]
    \centering
    \includegraphics[width=0.75\linewidth]{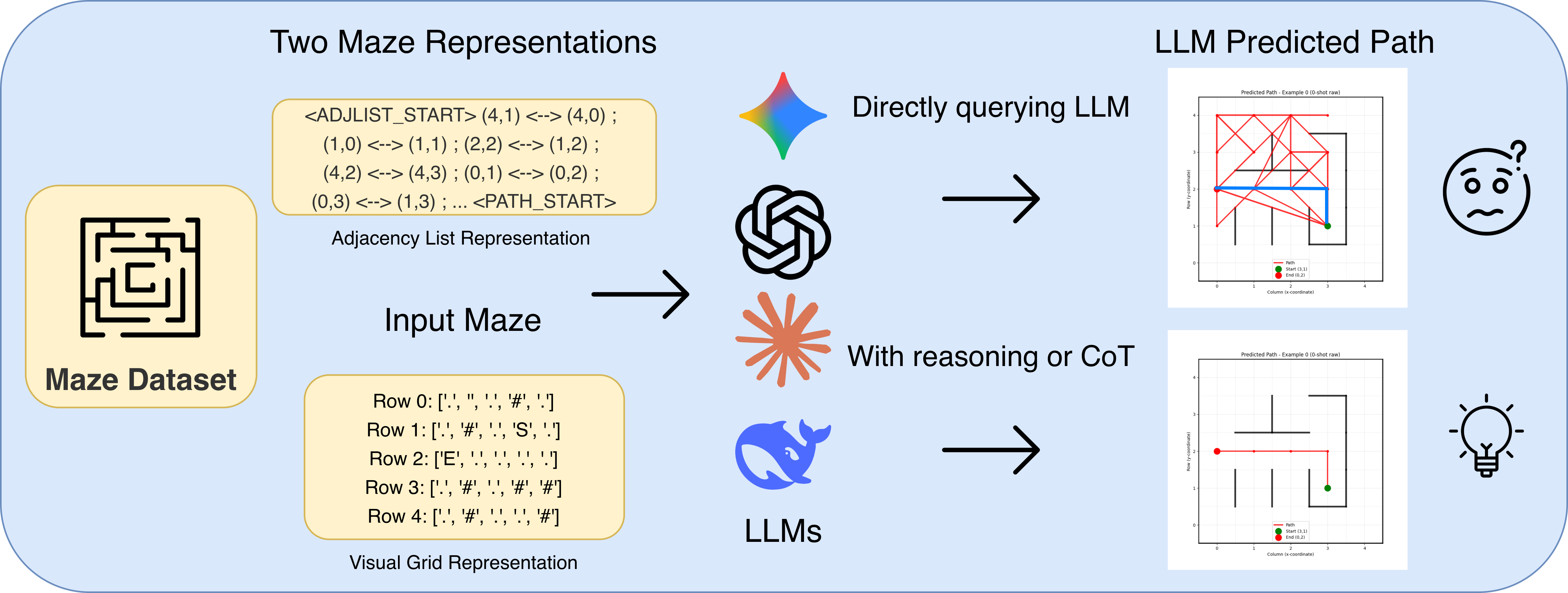}
    \caption{Overview of the main maze navigation task. This illustrates our systematic framework for evaluating whether LLMs construct spatial abstractions or rely on surface-level pattern matching. Models are asked to navigate mazes presented in two distinct formats: adjacency lists and visual character grids. Here, we present a comparison between direct querying LLMs, where models tend to fail the task by getting lost in the maze, and successfully solving the task through reasoning-based prompting. The latter can elicit more structured and valid navigation paths.}
    \label{fig:overview-fig}
\end{figure}

We systematically evaluated multiple LLM models: commercial foundation models (Gemini-2.5-Flash, GPT-5-mini, Claude-Haiku-4.5, and DeepSeek-Chat) accessed via API, and targeted spatial reasoning probes for proximity and directional understanding. We evaluate these models on generated mazes ranging from $5 \times 5$ to $9 \times 9$ grids, employing two complementary input representations: adjacency lists with special tokens (\texttt{<ADJLIST\_START>}, \texttt{<ORIGIN\_START>}) and human-readable visually formatted grids (Figure ~\ref{fig:overview-fig}). Through systematic manipulation of prompting strategies (standard, chain-of-thought, and explicit reasoning) and $k$-shot learning conditions (0, 3, and 5 examples), we isolate factors influencing spatial planning performance. Success metrics encompass exact path accuracy, edit distance, path validity, and reasoning quality analysis.

Our experiments reveal several critical patterns that enhance our understanding of spatial reasoning in LLMs. LLMs are sensitive to the input format of the maze, with a 2–5× accuracy drop when Gemini switches from tokenized to visual representations, despite an identical underlying maze structure. For example, Gemini-2.5-Flash with CoT achieves 86\% on 5x5 tokenized mazes but only 34\% on visual grid format. Moreover, without explicit reasoning instructions, most models fail on the task. For example, Claude-Haiku-4.5 fails almost completely on 5x5 tokenized mazes with base prompting, yet recovers to 78\% with CoT, demonstrating that spatial reasoning capabilities require clear instructions. Finally, models largely treat sequential spatial queries as independent questions and fail to leverage previously computed information even when identical questions are being asked. These findings collectively demonstrate that current LLMs exhibit representation-dependent, architecture-specific reasoning confined to narrow operational conditions rather than robust spatial abstraction.

Our primary contributions and observations are as follows: 
\begin{enumerate}[itemsep=0pt, topsep=0pt, parsep=0pt, partopsep=0pt]
    \item We present a systematic evaluation framework for probing spatial world models in foundation models, with manipulation of input format, model architecture, and prompting strategy.
    \item We provide empirical evidence characterizing conditions under which LLMs succeed and fail at spatial planning, revealing critical dependencies on representation and architecture.
    \item We offer methodological insights for the broader planning and foundation models community, including the demonstration that reasoning quality does not uniformly predict task success, and results reveal a dissociation between task accuracy and reasoning quality.
    \item Further, models fail to maintain spatial knowledge across sequential queries, treating repeated questions as independent events rather than leveraging previously computed information.
    \item Finally, we also built a platform for evaluating spatial world models \href{https://huggingface.co/spaces/weijiang99/SpatialBench}{SpatialBench} on HuggingFace Spaces.
\end{enumerate}

\section{Related Work}
\subsection{Maze Solving with LLMs}

Maze navigation has emerged as a context for studying spatial reasoning and world model formation. \cite{nolte2024transformers} demonstrates that transformers trained on maze-solving tasks acquire linear representations capturing maze connectivity encoded within single token latent states, with embedding layers learning spatial structure and specialized ``adjacency heads" respecting maze topology. \cite{spies2024transformers} extends this analysis using sparse autoencoders, further showing that transformers construct causal world models for maze tasks and that models can reason with respect to more active features than those encountered during training. These interpretability studies employ the maze-dataset library \citep{ivanitskiy2025maze}, which provides configurable maze generation algorithms and tokenization methods designed for autoregressive transformers. Additionally, \cite{dao2025alphamaze} explores LLM querying strategies and visual representations for maze-solving tasks, though the performance remains limited under complex experiment settings. 


\subsection{Model Planning Abilities and World Models}
With the emergence of different types of transformer-based models, more studies are being done on probing models' graph-based planning capabilities. For example, \citep{lehnert2024beyond} studies transformer planning capabilities and obtained a Searchformer for solving Sokoban puzzles with high accuracy. Further, \cite{tang2024grapharena} shows that even top-performing LLMs still struggle with more complex graph problems and exhibit hallucination issues. ~\cite{wu2024can} studies found LLM's solutions lack invariance under graph isomorphism. ~\cite{cai2025logiplan} shows that significant performance gaps remain in relational reasoning tasks with graph structures.

\subsection{Representation Alignment and World Models in LLMs}

A growing body of work argues that robust reasoning and planning in foundation models depends critically on representation alignment—both alignment across different internal representations within a model, across models, and between model representations and human conceptual structures. ~\cite{beger2026aimodelsperformhumanlike} formalize this view by distinguishing surface-level heuristic competence from genuinely aligned internal world models, emphasizing that reliable generalization requires representations that preserve task-relevant structure across contexts and formats. Similarly, ~\cite{lewis2024evaluatingrobustnessanalogicalreasoning} argue that failures in reasoning often arise not from lack of information, but from misaligned internal representations that prevent models from consistently reusing previously inferred knowledge.

Recent work has begun to operationalize representation alignment in the context of Theory-of-Mind (ToM) and world-model tasks.~\cite{das2025representationalignment} demonstrates that models with similar aggregate performance can rely on qualitatively different internal representations, leading to divergent generalization behavior under distributional or representational shifts. This finding challenges the assumption that strong task accuracy implies shared or stable internal world models. Complementarily, ~\cite{sucholutsky2024gettingalignedrepresentationalalignment} proposes a framework for analyzing whether models encode task structure in a format-invariant manner, arguing that alignment should be evaluated through invariance tests across equivalent representations rather than through single-format benchmarks.

Our work directly extends this line of inquiry to spatial world models.

\subsection{Benchmarks for World Models}
Evaluating whether models construct genuine world models requires benchmarks that probe internal structure, persistence, and generalization, rather than surface-level task success. However, recent critiques highlight that many benchmarks conflate task performance with world-model formation, allowing models to succeed via shallow pattern matching or shortcut learning.   To evaluate how LLMs can reason, plan and make decisions in spatial and non-spatial tasks, ~\cite{kokel2025acpbench} have proposed a benchmark to automatically synthesize problems with provably correct solutions across many types of tasks.  This work adopts an analogous approach to synthetically generate different types and sizes of grid-world maze task to systematically study spatial abstraction in LLMs.


\section{Method}
\subsection{Problem Definition}

We evaluate language models on a set of maze-solving tasks that require spatial reasoning and planning in grid-based environments. The tasks can be classified as planning or reasoning. In a planning task, the language model generates a sequence of actions that transforms an initial state to a goal state, e.g., the shortest path from A to B in the maze. In a reasoning task, the language model infers over representations to drive logical conclusions, e.g., is A closer to B than C in the maze.

\paragraph{Task 1: Maze Navigation (Planning)} Given a maze represented as an $n \times n$ grid of cells with specified connections between adjacent cells, the model must find a valid path from the start to the goal. \textbf{Input:} We consider two input formats: one is \textbf{adjacency list format}, a textual representation listing all connections between adjacent cells (e.g., ``$(0,0) <--> (0,1)$''), along with marked start and target positions; the other is \textbf{visual grid format}, which is a grid format that explicitly shows open paths, walls, start point, and end point in the maze. \textbf{Output:} The model generates a sequence of cell coordinates representing the path from start to target. A valid output must form a connected path where each consecutive pair of cells shares an edge in the maze.

\paragraph{Task 2: Sequential Reasoning with Point Reuse (Reasoning)} This task evaluates whether models can reuse previously acquired information. \textbf{Input:} Models are presented with a sequence of four proximity questions ($Q_0$, $Q_1$, $ Q_2$, $Q_3$) about the same maze. Each question requires the model to compare the relative distances between three distinct points within the grid. Critically, $Q_3$ is exactly the same as $Q_0$, as we want to establish a controlled probe for information reuse. By re-evaluating an identical spatial relationship after intermediate reasoning steps, we can determine whether the model leverages its previously computed reasoning or treats the repeated query as an isolated, independent question. \textbf{Output:} The model would answer four questions in order, outputting True or False along with its rationale as the reasoning. 

\paragraph{Task 3: Compositional Distance Comparison (Reasoning)} This task probes whether models can decompose complex spatial problems by reusing previously computed information. \textbf{Input:} The model is asked three questions regarding the relative distances between the maze corners ($A, B, C, D$) and the center ($M$). Given corners 
$A$ (top-left), $B$ (top-right), $C$ (bottom-left), and $D$ (bottom-right), 
and a center point $M$, we ask three proximity comparison questions in a \textit{single} prompt:
\begin{itemize}[noitemsep, topsep=0pt]
    \item $Q_0$: Is $A$ (top-left) closer to $M$ (center) than to $B$ (top-right)?
    \item $Q_1$: Is $D$ (bottom-right) closer to $M$ (center) than to $C$ (bottom-left)?
    \item $Q_2$: Is $B$ (top-right) closer to $C$ (bottom-left) than to $M$ (center)?
\end{itemize} 

\textbf{Output:} The model would answer four questions in order, answering with True or False along with its reasoning for the answer.

\subsection{Maze Dataset and Maze Representation}
\textbf{Dataset Construction} Mazes were generated using the Maze Dataset 
\cite{ivanitskiy2025maze}, employing depth-first search with percolation 
($p=0.2$) to create navigation challenges featuring cycles and multiple 
viable paths. For each grid size $n \in \{5,6,7,8,9\}$, we generated 50 
test mazes with guaranteed start-to-goal connectivity. Start and goal 
positions were placed at opposite corners of the grid.

\textbf{Maze Input Format} Each maze was presented in two representation formats:

\noindent
\textbf{Adjacency List Format:} 
{\footnotesize
\begin{verbatim}
<ADJLIST_START> (0,0) <--> (0,1) (0,1) <--> (1,1) <ADJLIST_END>
<ORIGIN_START> (0,0) <TARGET_START> (4,4)  
<PATH_START> (0,0) (0,1) (1,1) (2,1) (4,4) <PATH_END>
\end{verbatim}
}
This format encodes maze topology as an explicit adjacency list with special delimiter tokens, providing unambiguous structural information.

\noindent
\textbf{Visual Grid Format:} 
{\footnotesize
\begin{verbatim}
Row 0: ['.', 'S', '.', '#'] 
Row 1: ['#', '.', '.', '.']
Row 2: ['#', '#', '.', '#']
Row 3: ['.', '.', '#', 'E'] 
Legend: '#' = walls, '.' = open paths, 'S' = start, 'E' = end
\end{verbatim}
}
This format presents the maze as a human-readable 2D character array with explicit legend and coordinate system explanation, testing whether models can leverage visual-spatial intuitions from pre-training.


\subsection{Querying LLMs for Maze-Solving Tasks}

To investigate whether LLMs develop internal world models for spatial reasoning and planning, we employed three prompting strategies that vary in when and how reasoning is elicited:

\textbf{Standard Prompting} Zero-shot and few-shot ($k \in \{0, 3, 5\}$) prompts with task instructions and output format specification. Models receive the maze representation and are asked to output the solution path as a coordinate sequence: \texttt{(0,0) (0,1) (1,1) ...}. This baseline tests whether models can solve mazes without explicit reasoning instructions.

\textbf{Chain-of-Thought (CoT)} Models are instructed to reason step-by-step 
\emph{before} producing the coordinate path. The prompt guides models through a structured reasoning process:
\textit{``Before providing your final answer, think step-by-step about how to solve this maze: (1) First, analyze the maze structure and identify the start and end positions (2) Look for obvious paths and potential obstacles (3) Consider different possible routes (4) Choose the shortest valid path (5) Verify that your path does not cross walls or go out of bounds''}
This approach tests whether explicit reasoning instructions would improve spatial understanding.

\textbf{Post-hoc Reasoning} Models first provide the coordinate path, then explain their spatial analysis. This tests whether models can articulate coherent reasoning \emph{after} producing an answer, potentially revealing implicit world model usage:
\textit{`After providing your coordinate path, please explain your reasoning step-by-step: (1) How did you analyze the maze structure? (2) What strategy did you use to find the path? (3) Why did you choose this specific path? (4) What obstacles or dead ends did you avoid? (5) How did you ensure the path is valid and does not cross walls?''}





\subsection{Evaluation Metrics}
\subsubsection{Maze-Solving Tasks Evaluation Metrics}

For maze-solving tasks, we measure model performance using accuracy, defined by the proportion of generated paths that exactly match the ground truth solution or models correctly answering the proximity comparison questions. 

\subsubsection{Reasoning Quality Evaluation Metrics}

For reasoning tasks, we employed two metrics for evaluating the quality and consistency of LLM reasoning. Semantic coverage measures high-level conceptual similarity, and it is often sufficient for tasks that require language understanding. In our reasoning testing scenarios, we aim to determine not only if the responses are conceptually related but also whether the model is actively reusing specific information, so we employ ROUGE-L for counting exact text matches and LLM-as-a-judge to thoroughly compare two reasoning texts. 

Semantic coverage chain employed from a previous work \cite{golovneva2022roscoe}. Semantic coverage measures the overall degree of similarity between the reference and hypothesis chains, and is defined by the following: suppose the reasoning for the previous question is $r_0$, and the reasoning for the current question is $r_1$, $SC = [1 + cos(emb(r_0), emb(r_1))]/2$, where $emb(r_0)$ and $emb(r_1)$ are the sentence embedding for $r_0$ and $r_1$, respectively, computed by SentenceTransformer \citep{reimers-2019-sentence-bert}. 

Also, we measure the ROUGE-L score \citep{lin2004rouge} between $r_0$ and $r_1$ by finding the Longest Common Subsequence (LCS). This metric identifies the longest sequence of words appearing in both $r_0$ and $r_i$ in their original relative order. It is calculated based on the LCS between $r_0$ and $r_1$, which allows it to capture sentence-level structure similarity more flexibly than strict keyword matching. 

Semantic coverage and ROUGE-L score measure how similar two pieces of reasoning text are, the more similar indicating that more information is overlapped between $r_0$ and $r_1$. However, it is a quantitative metric, regardless of the actual content of the text. Thus, for reasoning tasks (Task 2 and Task 3), we also leverage  LLM-as-a-judge \cite{gu2025surveyllmasajudge}, evaluating the information reuse with an LLM(Llama-3.1-8b). The LLM is instructed to determine how much useful information acquired in the previous question for solving the maze-related question, such as calculations or spatial information, is being included in the later question. The full prompt is included in Section ~\ref{sec:llm-as-judge-prompt}.

\section{Results}

\begin{figure}
    \centering
    \includegraphics[width=0.5\linewidth]{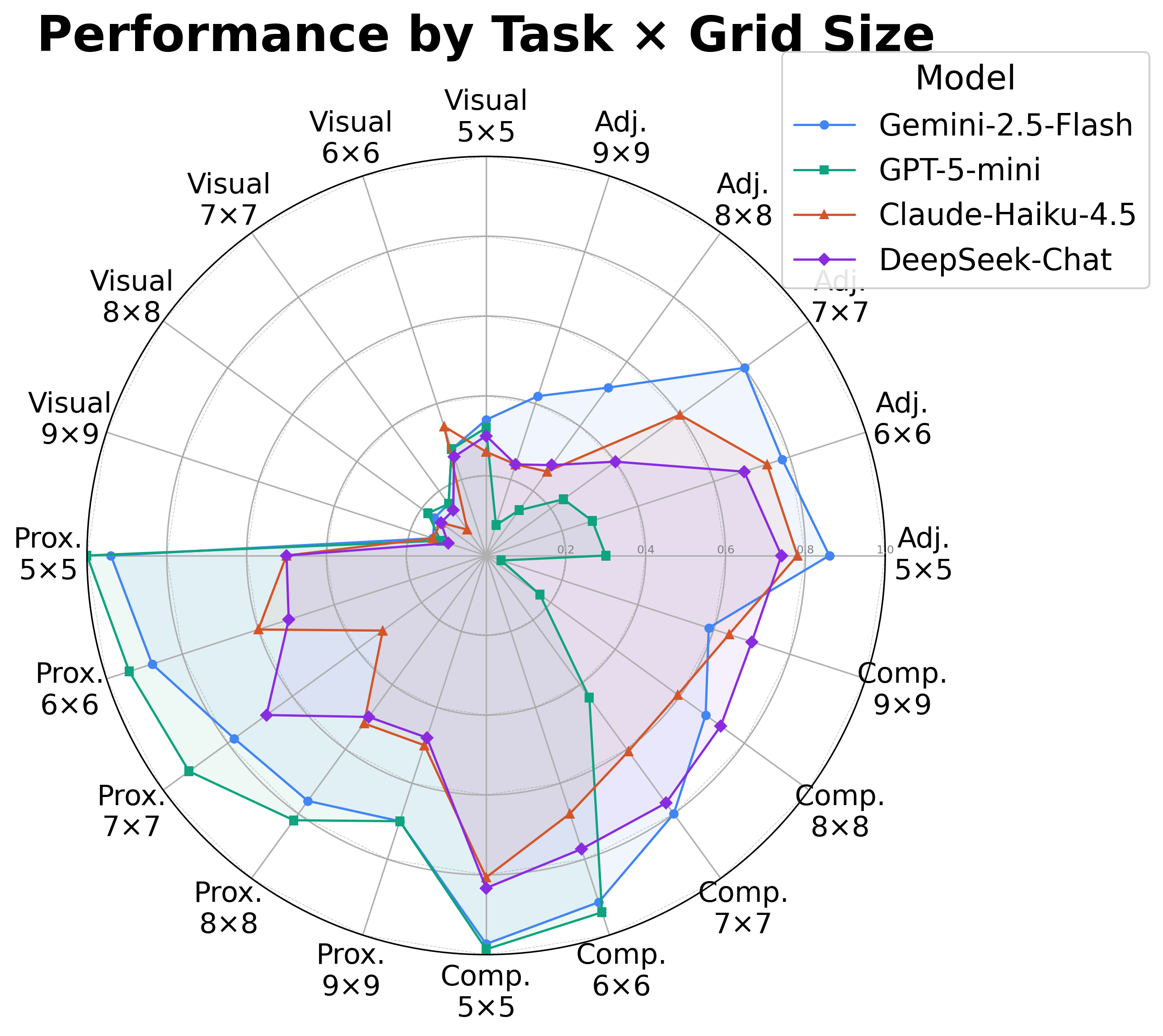}
    \caption{An overview of model performance across all tasks and maze sizes, with each spoke representing a maze-solving task and grid size, and colored lines representing the average performance of the LLM over all mazes generated for that task.}
    \label{fig:results-spider-plot}
\end{figure}

We evaluated 4 LLMs on three main tasks, including maze navigation, sequential distance comparison, and compositional distance comparison. Figure \ref{fig:results-spider-plot} summarizes model performance across all three tasks and five maze sizes. Several patterns emerge: first, performance consistently degrades as maze size increases from $5 \times 5$ to $9 \times 9$ across all tasks and models; second, adjacency list formats substantially outperform visual grid formats; third, different models excel at different tasks; finally, the inner region of the chart, corresponding to larger mazes, remains largely unoccupied, highlighting the limits of current LLMs on complex spatial planning. The following subsections present results for each task in detail. 

\subsection{Maze Representation and $k$-Shot Learning}

We evaluate four LLMs on the maze-solving task, where models are asked to generate the shortest path between designated start and end points given a maze representation. Each model receives the maze structure in one of two formats, adjacency lists or visual grids, and is asked to output a sequence of coordinates forming a valid path. We test three prompting strategies: standard (base) prompting with only task instructions, chain-of-thought (CoT) prompting that instructs models to reason step-by-step before answering, and post-hoc reasoning where models provide the path first, then explain their approach. For each configuration, we evaluate on 50 mazes per grid size ($5 \times 5$ through $9 \times 9$).

Table ~\ref{tab:experiment-1-results} in the Section \ref{sec:appendix} presents the results for four models on zero-shot maze-solving questions. Here, we show how different sizes of LLM models perform on maze-solving tasks with different sizes of the maze. Larger mazes are usually more challenging to solve; all models have a performance decline with larger mazes, with results for 3-shot and 5-shot included in Table ~\ref{tab:3-shot-learning-results} and Table ~\ref{tab:5-shot-learning-results} in Section ~\ref{sec:appendix}. Further, we observed the following patterns:

\textbf{Prompting strategy is critical for spatial reasoning.} Without explicit reasoning instructions, most models fail on both representations of the maze, especially for Claude-Haiku-4.5 and DeepSeek-Chat models, with poor performance with the base prompt. Visualizations of the model-generated paths demonstrate that the model tends to end up getting lost in the maze after finding the shortest Euclidean distance, ignoring the maze grid structure and wall obstacles, as this is the most common failure mode. However, all models show significant improvements when explicitly instructed to recognize the maze structure and reflect on the path-finding process. For example, Claude-Haiku-4.5's performance recovers to an accuracy of 78\% with the chain-of-thought instruction, and similar patterns exist in DeepSeek-Chat, demonstrating that spatial reasoning capabilities exist but require appropriate elicitation. 

\textbf{Adjacency list representations outperform visual formats in most cases.} When using CoT or reasoning prompts, 3/4 models achieve substantially higher accuracy on adjacency lists compared to visual character grids. For example, Gemini-2.5-Flash with CoT reaches 86\% on 5×5 adjacency list mazes but only 34\% on the visual grid equivalent. This gap persists across grid sizes: on 7×7 mazes, Gemini achieves 80\% with adjacency list representation, compared to 16\% with the visual representation. The consistent advantage of adjacency list formats suggests that explicit structural information, such as adjacency lists, is more amenable to LLM processing than spatial layouts requiring visual parsing. Only GPT-5-mini's performance did not conform to this observation. 

\subsection{Sequential Questions with Point Reuse}

As we started the exploration with LLM k-shot learning on the maze-solving task, we noticed that the \textit{adjacency list} representation yields a better performance. Thus, we use \textit{adjacency list} representation for the following experiments. Next, we are interested in investigating whether LLM understands the structure of the maze while solving the maze, or if it obtains the path via brute-force search. 

We evaluated four models on 50 mazes per grid size (5×5 to 9×9), presenting sequences of four proximity comparison questions ($Q_0$ through $Q_3$) within a single prompt. Specifically, $Q_3$ repeated the same query as $Q_0$, which guides us to understand and assess the reasoning consistency and potential information reuse among LLMs. Table ~\ref{tab:experiment-2-results} shows the per-question accuracy across four models. 

\subsubsection{Reasoning Quality}

To evaluate whether LLMs maintain consistent spatial representations or treat repeated questions as independent tasks, we analyzed the reasoning traces between $Q_0$ and $Q_3$ using semantic coverage (SC), ROUGE-L (R-L), and LLM-as-a-judge metrics. As shown in Table ~\ref{tab:experiment-2-reasoning-quality}, all models exhibit relatively high SC scores, suggesting high conceptual similarity. However, the literal overlap measured by ROUGE-L reveals different internal behaviors among different models. 

\textbf{Re-calculation of the same question.} Gemini-2.5-Flash and GPT-5-mini maintain remarkably high ROUGE-L scores, indicating a tendency to treat $Q_3$ independently by re-executing its full problem-solving process. Qualitatively, Gemini-2.5-Flash consistently reproduces detailed, step-by-step Breadth-First Search (BFS) traversals for both questions. On the other hand, GPT-5-mini also treats each question individually, but it directly shows the shortest path between two points and then compares the distance. This re-computation ensures high accuracy across repeated questions, but suggests the model does not explicitly memoize or reference the prior computation. 

\textbf{Minimal Reasoning for Repeated Question.} In contrast, DeepSeek-Chat demonstrates significantly lower ROUGE-L scores, with the lowest being 0.08, despite high SC scores, as other models also have high SC scores across all settings. This pattern stems from minimal reasoning in the repeated question, as the model saved the detailed process and only outputs the final answer. 

\textbf{Explicit Signals for Information Reuse.} Similarly, Claude demonstrates similar patterns to DeepSeek, as both have significantly lower ROUGE-L scores compared to the other two models. In Claude's response in $Q_3$, it would explicitly mention that ``From my previous analysis'' and reuses the computed distances from previous questions, or it would acknowledge $Q_3$ being the same as $Q_0$ clearly in its answer, and thus the generated answers and intuition are much shorter than $Q_0$. 

\vspace{-2mm}
\begin{table}[h!]
\centering
\caption{Reasoning quality results for sequential questions with point reuse, measured by semantic coverage (SC), ROUGE-L (R-L), and LLM-as-a-judge (LLM)}
\label{tab:experiment-2-reasoning-quality}
\resizebox{\textwidth}{!}{%
\begin{tabular}{l|ccc|ccc|ccc|ccc|ccc}
\toprule
& \multicolumn{3}{c|}{\textbf{5×5}} 
& \multicolumn{3}{c|}{\textbf{6×6}} 
& \multicolumn{3}{c|}{\textbf{7×7}} 
& \multicolumn{3}{c|}{\textbf{8×8}} 
& \multicolumn{3}{c}{\textbf{9×9}} \\
\textbf{Model} 
& SC & R-L & LLM 
& SC & R-L & LLM 
& SC & R-L & LLM 
& SC & R-L & LLM 
& SC & R-L & LLM \\
\midrule
Gemini-2.5-Flash 
& 0.96 & 0.85 & \textbf{96.67} 
& 0.94 & 0.77 & \textbf{93.23} 
& 0.95 & \textbf{0.75} & \textbf{94.17} 
& 0.92 & \textbf{0.67} & \textbf{94.13} 
& 0.91 & 0.57 & \textbf{90.43} \\

GPT-5-mini 
& \textbf{0.98} & \textbf{0.89} & 93.47
& \textbf{0.98} & \textbf{0.87} & 92.19 
& \textbf{0.97} & 0.74 & 90.23 
& \textbf{0.93} & 0.52 & 85.00 
& \textbf{0.94} & \textbf{0.58} & 82.48 \\

Claude-Haiku-4.5 
& 0.86 & 0.40 & 85.90 
& 0.85 & 0.39 & 87.50 
& 0.85 & 0.37 & 86.53 
& 0.86 & 0.39 & 85.71 
& 0.87 & 0.36 & 85.0 \\

DeepSeek-Chat 
& 0.86 & 0.27 & 87.40 
& 0.86 & 0.26 & 87.04 
& 0.86 & 0.23 & 86.84 
& 0.81 & 0.10 & 88.33 
& 0.82 & 0.08 & 87.62 \\

\bottomrule
\end{tabular}%
}
\end{table}
\vspace{-5mm}

\subsection{Compositional Distance Comparison}

Since the previous experiment setting did not explicitly force the LLMs to be aware of the spatial structure information, here we designed a multi-question task using the corner points and the center of each maze. To investigate whether LLMs can decompose complex spatial reasoning problems and reuse intermediate computations, the LLMs are given corners $A$ (top-left), $B$ (top-right), $C$ (bottom-left), and $D$ (bottom-right), and center point $M$, we asked three proximity comparison questions for 50 mazes within a \textit{single} prompt: $Q_0$ asks whether $A$ is closer to $M$ than to $B$; $Q_1$ asks whether $D$ is closer to $M$ than to $C$; and $Q_2$ asks whether $B$ is closer to $C$ than to $M$. Crucially, $Q_2$ can potentially benefit from composing the corner-to-center distance information established in $Q_0$ and $Q_1$, since paths between opposite corners often traverse the center region. The prompt explicitly encouraged models to reuse information across questions.

Results for this task are presented in Table ~\ref{tab:experiment-3-results}, showing the per-question accuracy across four models. Overall performance ranged from 51.1\% to 79.1\%, with all models performing well above a random guess (50\%). However, the key measuring metric here is $Q_2$'s accuracy relative to the average accuracy of $Q_0$ and $Q_1$. A positive difference would suggest the model benefits from the decomposition of the question, indicating that the model leverages distance previously computed. We observed that the Gemini and Claude models show slight improvements in $Q_2$, whereas DeepSeek and GPT-5-mini showed a decline in performance, which might be due to the extended context length or the complexity of the question.

\subsubsection{Reasoning Quality}

To assess whether models genuinely compose intermediate results across questions, we evaluate reasoning quality using three complementary metrics: semantic coverage, ROUGE-L, and LLM-as-a-judge. 



\textbf{Hints for Previously Acquired Information} In this task, Gemini shows hints for information re-use in $Q_3$ for some of the questions. For example, in Gemini's generated responses, it would explicitly mention ``From the previous BFS, we already found this'' or  ``already found in the previous calculation'' in its answer, indicating what the model knows and which piece of information is helping it to solve $Q_3$. 

\textbf{Failure of Reusing Information.} GPT-5-mini, Claude-Haiku-4.5, and DeepSeek-Chat have similar behavior in this task setting, and all three models treat each question as a self-contained question. Specifically, for each question, the answer is formatted or framed in the same way, and the thinking process is mostly the same, as supported by the high semantic coverage scores across all models. We observed that in their generated responses, neither model mentioned any of the previously calculated results or visited paths. 

\vspace{-5mm}
\begin{table}[h!]
\centering
\caption{Reasoning quality results for compositional distance comparison, measured by semantic coverage (SC), ROUGE-L (R-L), and LLM-as-a-judge (LLM)}
\label{tab:experiment-3-reasoning-quality}
\resizebox{\textwidth}{!}{%
\begin{tabular}{l|ccc|ccc|ccc|ccc|ccc}
\toprule
& \multicolumn{3}{c|}{\textbf{5×5}} 
& \multicolumn{3}{c|}{\textbf{6×6}} 
& \multicolumn{3}{c|}{\textbf{7×7}} 
& \multicolumn{3}{c|}{\textbf{8×8}} 
& \multicolumn{3}{c}{\textbf{9×9}} \\
\textbf{Model} 
& SC & R-L & LLM 
& SC & R-L & LLM 
& SC & R-L & LLM 
& SC & R-L & LLM 
& SC & R-L & LLM \\
\midrule
Gemini-2.5-Flash 
& 0.93 & 0.42 & 9.49 
& 0.92 & 0.39 & 16.25 
& 0.95 & 0.52 & \textbf{17.86} 
& 0.95 & 0.51 & \textbf{14.32} 
& 0.95 & 0.56 & \textbf{12.36} \\

GPT-5-mini 
& 0.95 & \textbf{0.59} & \textbf{23.4} 
& 0.95 & 0.60 & \textbf{17.1} 
& \textbf{0.96} & \textbf{0.82} & 16.1 
& \textbf{0.96} & \textbf{0.94} & 0.0 
& \textbf{0.98} & \textbf{0.99} & 0.0 \\

Claude-Haiku-4.5 
& 0.94 & 0.55 & 8.16 
& \textbf{0.96} & \textbf{0.62} & 7.0 
& 0.94 & 0.54 & 10.0 
& 0.95 & 0.55 & 4.6 
& 0.95 & 0.59 & 13.0 \\

DeepSeek-Chat 
& \textbf{0.96} & \textbf{0.59} & 3.0 
& \textbf{0.96} & 0.60 & 0.0 
& 0.94 & 0.56 & 5.0 
& 0.95 & 0.55 & 5.6 
& 0.93 & 0.50 & 10.9 \\

\bottomrule
\end{tabular}%
}
\end{table}

\section{Discussion and Limitation}
In this work, we demonstrate that while LLMs like Gemini-2.5-Flash and Claude-Haiku-4.5 can achieve high accuracy on maze navigation tasks, this success is limited to the adjacency list representation of the maze, suggesting that models rely on representation-specific heuristics rather than abstract spatial understanding. Furthermore, our analysis of reasoning quality using semantic coverage, ROUGE-L, and LLM-as-a-judge reveals that models tend to treat spatial queries as independent events. Even when presented with identical or compositional questions, models tend to work on the question from scratch rather than leveraging cumulative information. 

We note that our conclusions are drawn from behavioral evidence rather than from direct analysis of internal representations. Models may encode spatial structure internally but fail to leverage it consistently under certain prompts or representations. Additionally, both input formats in our experiments remain text-based. While this controlled design isolates representational effects from perceptual ones, it leaves open the question of how these findings transfer to settings involving actual visual perception or 3D spatial reasoning.

Our proposed evaluation framework, involving synthetic task generation with provably correct solutions, explicitly disentangles spatial reasoning from spatial planning, and systematically varies representation format, prompting strategy, and task composition. By combining maze navigation with sequential proximity queries and compositional distance comparisons, we directly test whether models build persistent, format-invariant spatial world models. The consistent failure of models to generalize across representations or reuse spatial knowledge positions our benchmark as a diagnostic tool for identifying representation-dependent reasoning, a critical limitation of current foundation models. To facilitate reproducibility, we release SpatialBench, an open evaluation platform on HuggingFace Spaces\footnote{\url{https://huggingface.co/spaces/weijiang99/SpatialBench}} with an interactive leaderboard and experiment script generator.

\subsection{Limitations and Future Work}
While our framework offers a context for probing spatial reasoning, several limitations remain. First, the scope of environments is limited to 2D grid mazes, and future work should explore more complex, non-Euclidean graphs to test the limits of spatial abstraction. Secondly, we noted that $k$-shot learning often degrades performance on larger mazes or with longer context length, with more examples provided. Investigating the mechanism behind this performance gap is crucial. Finally, while we characterized model behavioral patterns, we did not investigate the internal representations that drive the model's performance. Future work employing probing classifiers, activation analysis, or interpretability methods could reveal whether models encode any spatial structure internally.

\bibliography{iclr2026_conference}
\bibliographystyle{iclr2026_conference}

\appendix
\section{Appendix}\label{sec:appendix}
\subsection{Additional maze-solving results with fine-tuned T5 models}

Initially, we followed the same idea proposed in \citep{ivanitskiy2023structured} that uses a transformer-based encoder-decoder model and fine-tuned T5 models using data generated from the maze-dataset. 

\begin{table}[h!]
\centering
\caption{Fine-tuned model performance on different grid sizes (Path Accuracy). Each model is trained with 8000 examples and tested with 2000 examples over 10 epochs.}
\begin{tabular}{c|ccccc}
\hline
\multirow{2}{*}{\textbf{Train Grid Size}} & \multicolumn{5}{c}{\textbf{Test Grid Size}} \\
\cline{2-6}
 & \textbf{5×5} & \textbf{6×6} & \textbf{7×7} & \textbf{8×8} & \textbf{9×9} \\
\hline
5×5 & 0.806 & 0.407 & 0.251 & 0.152 & 0.093 \\
6×6 & 0.856 & 0.769 & 0.449 & 0.258 & 0.168 \\
7×7 & 0.820 & 0.746 & 0.691 & 0.400 & 0.248 \\
8×8 & 0.799 & 0.718 & 0.673 & 0.614 & 0.373 \\
9×9 & 0.778 & 0.694 & 0.623 & 0.567 & 0.537 \\
\hline
\end{tabular}
\label{tab:fine-tuned-t5-model-performance}
\end{table}

\subsection{$k$-shot learning results}
\begin{table}[h!]
\centering
\caption{Model performance on zero-shot maze solving task, results are reported as accuracy over 50 mazes}
\label{tab:experiment-1-results}
\begin{adjustbox}{max width=\textwidth}
\begin{tabular}{ll*{5}{S[table-format=2.1]S[table-format=2.1]}}
\toprule
& & \multicolumn{2}{c}{\textbf{5×5}} & \multicolumn{2}{c}{\textbf{6×6}} & \multicolumn{2}{c}{\textbf{7×7}} & \multicolumn{2}{c}{\textbf{8×8}} & \multicolumn{2}{c}{\textbf{9×9}} \\
\cmidrule(lr){3-4} \cmidrule(lr){5-6} \cmidrule(lr){7-8} \cmidrule(lr){9-10} \cmidrule(lr){11-12}
\textbf{Model} & \textbf{Setting} & {Adj} & {Visual} & {Adj} & {Visual} & {Adj} & {Visual} & {Adj} & {Visual} & {Adj} & {Visual} \\
\midrule
\multirow{3}{*}{Gemini-2.5-Flash} 
    & Base      & 0.08 & 0.34 & 0.06 & 0.34 & 0.06 & 0.16 & 0.08 & 0.14 & 0.02 & 0.10 \\
    & CoT       & 0.86 & 0.34 & 0.78 & 0.28 & 0.80 & 0.16 & 0.52 & 0.16 & 0.42 & 0.14 \\
    & Reasoning & 0.78 & 0.30 & 0.84 & 0.30 & 0.72 & 0.16 & 0.38 & 0.14 & 0.36 & 0.10 \\
\midrule
\multirow{3}{*}{GPT-5-mini} 
    & Base      & 0.32 & 0.32 & 0.20 & 0.32 & 0.22 & 0.16 & 0.14 & 0.16 & 0.10 & 0.10 \\
    & CoT       & 0.30 & 0.32 & 0.28 & 0.28 & 0.24 & 0.16 & 0.14 & 0.18 & 0.08 & 0.12 \\
    & Reasoning & 0.16 & 0.38 & 0.10 & 0.30 & 0.12 & 0.14 & 0.12 & 0.20 & 0.06 & 0.10 \\
\midrule
\multirow{3}{*}{Claude-Haiku-4.5} 
    & Base      & 0.00 & 0.00 & 0.28 & 0.00 & 0.00 & 0.00 & 0.00 & 0.00 & 0.00 & 0.00 \\
    & CoT       & 0.78 & 0.26 & 0.74 & 0.34 & 0.60 & 0.08 & 0.26 & 0.14 & 0.24 & 0.14 \\
    & Reasoning & 0.74 & 0.32 & 0.66 & 0.24 & 0.42 & 0.14 & 0.26 & 0.12 & 0.24 & 0.10 \\
\midrule
\multirow{3}{*}{DeepSeek-Chat} 
    & Base      & 0.14 & 0.16 & 0.06 & 0.10 & 0.00 & 0.06 & 0.00 & 0.06 & 0.00 & 0.08 \\
    & CoT       & 0.74 & 0.30 & 0.68 & 0.26 & 0.40 & 0.14 & 0.28 & 0.14 & 0.24 & 0.10 \\
    & Reasoning & 0.80 & 0.12 & 0.54 & 0.08 & 0.44 & 0.06 & 0.26 & 0.06 & 0.26 & 0.08 \\
\bottomrule
\end{tabular}
\end{adjustbox}
\end{table}

Additional results for $k$-shot learning on maze-solving tasks with $k = {3, 5}$, evaluated on 50 mazes for each grid size

\begin{table}[h!]
\centering
\caption{Model performance on 3-shot maze solving task, results are reported as accuracy}
\label{tab:3-shot-learning-results}
\begin{adjustbox}{max width=\textwidth}
\begin{tabular}{ll*{5}{S[table-format=2.1]S[table-format=2.1]}}
\toprule
& & \multicolumn{2}{c}{\textbf{5×5}} & \multicolumn{2}{c}{\textbf{6×6}} & \multicolumn{2}{c}{\textbf{7×7}} & \multicolumn{2}{c}{\textbf{8×8}} & \multicolumn{2}{c}{\textbf{9×9}} \\
\cmidrule(lr){3-4} \cmidrule(lr){5-6} \cmidrule(lr){7-8} \cmidrule(lr){9-10} \cmidrule(lr){11-12}
\textbf{Model} & \textbf{Setting} & {Adj} & {Visual} & {Adj} & {Visual} & {Adj} & {Visual} & {Adj} & {Visual} & {Adj} & {Visual} \\
\midrule
\multirow{3}{*}{Gemini-2.5-Flash} 
    & Base      & 0.32 & 0.32 & 0.26 & 0.3 & 0.36 & 0.16 & 0.2 & 0.12 & 0.1 & 0.1 \\
    & CoT       & 0.82 & 0.32 & 0.84 & 0.30 & 0.74 & 0.22 & 0.68 & 0.12 & 0.44 & 0.10 \\
    & Reasoning & 0.82 & 0.30 & 0.78 & 0.32 & 0.62 & 0.16 & 0.56 & 0.14 & 0.36 & 0.08 \\
\midrule
\multirow{3}{*}{GPT-5-mini} 
    & Base      & 0.42 & 0.36 & 0.38 & 0.38 & 0.16 & 0.16 & 0.18 & 0.20 & 0.06 & 0.08 \\
    & CoT       & 0.34 & 0.32 & 0.24 & 0.28 & 0.12 & 0.16 & 0.12 & 0.14 & 0.06 & 0.10 \\
    & Reasoning & 0.28 & 0.36 & 0.22 & 0.28 & 0.16 & 0.18 & 0.12 & 0.16 & 0.04 & 0.08 \\
\midrule
\multirow{3}{*}{Claude-Haiku-4.5} 
    & Base      & 0.00 & 0.10 & 0.28 & 0.12 & 0.00 & 0.06 & 0.00 & 0.00 & 0.00 & 0.00 \\
    & CoT       & 0.74 & 0.26 & 0.74 & 0.24 & 0.58 & 0.14 & 0.44 & 0.16 & 0.26 & 0.14 \\
    & Reasoning & 0.78 & 0.34 & 0.68 & 0.28 & 0.42 & 0.18 & 0.40 & 0.12 & 0.30 & 0.10 \\
\midrule
\multirow{3}{*}{DeepSeek-Chat} 
    & Base      & 0.48 & 0.28 & 0.28 & 0.22 & 0.28 & 0.00 & 0.36 & 0.12 & 0.20 & 0.06 \\
    & CoT       & 0.68 & 0.30 & 0.44 & 0.24 & 0.28 & 0.14 & 0.24 & 0.12 & 0.24 & 0.10 \\
    & Reasoning & 0.44 & 0.16 & 0.24 & 0.22 & 0.20 & 0.14 & 0.36 & 0.10 & 0.16 & 0.08 \\
\bottomrule
\end{tabular}
\end{adjustbox}
\end{table}

\begin{table}[h!]
\centering
\caption{Model performance on 5-shot maze solving task, results are reported as accuracy}
\label{tab:5-shot-learning-results}
\begin{adjustbox}{max width=\textwidth}
\begin{tabular}{ll*{5}{S[table-format=2.1]S[table-format=2.1]}}
\toprule
& & \multicolumn{2}{c}{\textbf{5×5}} & \multicolumn{2}{c}{\textbf{6×6}} & \multicolumn{2}{c}{\textbf{7×7}} & \multicolumn{2}{c}{\textbf{8×8}} & \multicolumn{2}{c}{\textbf{9×9}} \\
\cmidrule(lr){3-4} \cmidrule(lr){5-6} \cmidrule(lr){7-8} \cmidrule(lr){9-10} \cmidrule(lr){11-12}
\textbf{Model} & \textbf{Setting} & {Adj} & {Visual} & {Adj} & {Visual} & {Adj} & {Visual} & {Adj} & {Visual} & {Adj} & {Visual} \\
\midrule
\multirow{3}{*}{Gemini-2.5-Flash} 
    & Base      & 0.24 & 0.30 & 0.14 & 0.26 & 0.36 & 0.20 & 0.20 & 0.14 & 0.10 & 0.14 \\
    & CoT       & 0.80 & 0.34 & 0.78 & 0.32 & 0.80 & 0.16 & 0.50 & 0.12 & 0.40 & 0.12 \\
    & Reasoning & 0.82 & 0.34 & 0.64 & 0.30 & 0.68 & 0.20 & 0.42 & 0.12 & 0.34 & 0.12 \\
\midrule
\multirow{3}{*}{GPT-5-mini} 
    & Base      & 0.40 & 0.38 & 0.28 & 0.24 & 0.28 & 0.14 & 0.16 & 0.16 & 0.08 & 0.10 \\
    & CoT       & 0.32 & 0.30 & 0.24 & 0.28 & 0.12 & 0.16 & 0.14 & 0.16 & 0.00 & 0.10 \\
    & Reasoning & 0.30 & 0.34 & 0.26 & 0.30 & 0.20 & 0.14 & 0.14 & 0.16 & 0.02 & 0.12 \\
\midrule
\multirow{3}{*}{Claude-Haiku-4.5} 
    & Base      & 0.00 & 0.08 & 0.28 & 0.10 & 0.00 & 0.06 & 0.00 & 0.00 & 0.00 & 0.02 \\
    & CoT       & 0.82 & 0.32 & 0.78 & 0.24 & 0.52 & 0.16 & 0.40 & 0.16 & 0.30 & 0.10 \\
    & Reasoning & 0.60 & 0.32 & 0.62 & 0.28 & 0.48 & 0.18 & 0.36 & 0.16 & 0.30 & 0.10 \\
\midrule
\multirow{3}{*}{DeepSeek-Chat} 
    & Base      & 0.32 & 0.16 & 0.32 & 0.12 & 0.32 & 0.00 & 0.16 & 0.04 & 0.00 & 0.00 \\
    & CoT       & 0.72 & 0.36 & 0.44 & 0.16 & 0.20 & 0.08 & 0.24 & 0.14 & 0.20 & 0.08 \\
    & Reasoning & 0.48 & 0.28 & 0.36 & 0.08 & 0.24 & 0.12 & 0.24 & 0.16 & 0.20 & 0.04 \\
\bottomrule
\end{tabular}
\end{adjustbox}
\end{table}

\newpage
\subsection{Example Prompt for Sequential Questions with Point Reuse}

\paragraph{Input Format Explanation}
\begin{itemize}[noitemsep, topsep=2pt]
    \item \textbf{Grid System:} Coordinates use $(\text{column}, \text{row})$ format, starting from $(0,0)$ at the top-left.
    \item \textbf{Adjacency List:} Each \texttt{(x,y) <--> (a,b)} indicates that cells $(x,y)$ and $(a,b)$ are directly connected.
    \item \textbf{Start Point:} \texttt{<ORIGIN\_START> (x,y)} indicates the starting position.
    \item \textbf{End Point:} \texttt{<TARGET\_START> (x,y)} indicates the destination.
    \item \textbf{Generation Cue:} \texttt{<PATH\_START>} signals the end of the maze definition.
\end{itemize}

\paragraph{Maze Input}
\begin{quote}\ttfamily
\texttt{<ADJLIST\_START>} (4,2) \texttt{<-->} (4,1) ; (3,4) \texttt{<-->} (2,4) ; (4,0) \texttt{<-->} (4,1) ;\\
(0,2) \texttt{<-->} (0,3) ; (0,0) \texttt{<-->} (0,1) ; (3,0) \texttt{<-->} (4,0) ; (2,2) \texttt{<-->} (3,2) ;\\
(4,2) \texttt{<-->} (4,3) ; (3,3) \texttt{<-->} (3,2) ; (3,0) \texttt{<-->} (2,0) ; (0,2) \texttt{<-->} (1,2) ;\\
(1,0) \texttt{<-->} (1,1) ; (4,4) \texttt{<-->} (3,4) ; (3,2) \texttt{<-->} (3,1) ; (0,4) \texttt{<-->} (1,4) ;\\
(2,3) \texttt{<-->} (2,4) ; (1,4) \texttt{<-->} (1,3) ; (0,2) \texttt{<-->} (0,1) ; (1,3) \texttt{<-->} (0,3) ;\\
(1,0) \texttt{<-->} (0,0) ; (1,2) \texttt{<-->} (2,2) ; (4,3) \texttt{<-->} (4,4) ; (2,1) \texttt{<-->} (2,0) ;\\
(2,1) \texttt{<-->} (2,2) ; \texttt{<ADJLIST\_END>} \texttt{<ORIGIN\_START>} (1,3) \texttt{<ORIGIN\_END>}\\
\texttt{<TARGET\_START>} (2,3) \texttt{<TARGET\_END>} \texttt{<PATH\_START>}
\end{quote}

\paragraph{Highlighted Positions}
A: $(4,3)$ \quad B: $(0,3)$ \quad C: $(1,3)$

\paragraph{General Instruction}
For each question, consider the \textbf{actual shortest path distance through the maze}, not straight-line (Euclidean) distance.

\paragraph{Answer Format}
\begin{quote}\ttfamily
True/False: [your explanation]
\end{quote}

\paragraph{Questions}
\begin{enumerate}
    \item \textbf{Q1.} In this maze, is position $(4,3)$ closer to position $(0,3)$ than it is to position $(1,3)$?

    \vspace{4pt}
    \textbf{Answer:}

    \item \textbf{Q2.} Next question about the same maze: Is position $(2,4)$ closer to position $(0,4)$ than it is to position $(2,3)$?

    \vspace{4pt}
    \textbf{Answer:}

    \item \textbf{Q3.} Next question about the same maze: Is position $(0,3)$ closer to position $(4,0)$ than it is to position $(1,0)$?

    \vspace{4pt}
    \textbf{Answer:}

    \item \textbf{Q4.} Next question about the same maze: Is position $(4,3)$ closer to position $(0,3)$ than it is to position $(1,3)$?

    \vspace{4pt}
    \textbf{Answer:}
\end{enumerate}

\subsection{Example Prompt for Compositional Distance Comparison}

\paragraph{Input Format Explanation}
\begin{itemize}[noitemsep, topsep=2pt]
    \item \textbf{Grid System:} Coordinates use $(\text{column}, \text{row})$ format, starting from $(0,0)$ at the top-left.
    \item \textbf{Adjacency List:} Each \texttt{(x,y) <--> (a,b)} indicates that cells $(x,y)$ and $(a,b)$ are directly connected.
    \item \textbf{Start Point:} \texttt{<ORIGIN\_START> (x,y)} indicates the starting position.
    \item \textbf{End Point:} \texttt{<TARGET\_START> (x,y)} indicates the destination.
    \item \textbf{Generation Cue:} \texttt{<PATH\_START>} signals the end of the maze definition.
\end{itemize}

\paragraph{Maze Input}
\begin{quote}\ttfamily
\texttt{<ADJLIST\_START>} (4,2) \texttt{<-->} (4,1) ; (3,4) \texttt{<-->} (2,4) ; (4,0) \texttt{<-->} (4,1) ;\\
(0,2) \texttt{<-->} (0,3) ; (0,0) \texttt{<-->} (0,1) ; (3,0) \texttt{<-->} (4,0) ; (2,2) \texttt{<-->} (3,2) ;\\
(4,2) \texttt{<-->} (4,3) ; (3,3) \texttt{<-->} (3,2) ; (3,0) \texttt{<-->} (2,0) ; (0,2) \texttt{<-->} (1,2) ;\\
(1,0) \texttt{<-->} (1,1) ; (4,4) \texttt{<-->} (3,4) ; (3,2) \texttt{<-->} (3,1) ; (0,4) \texttt{<-->} (1,4) ;\\
(2,3) \texttt{<-->} (2,4) ; (1,4) \texttt{<-->} (1,3) ; (0,2) \texttt{<-->} (0,1) ; (1,3) \texttt{<-->} (0,3) ;\\
(1,0) \texttt{<-->} (0,0) ; (1,2) \texttt{<-->} (2,2) ; (4,3) \texttt{<-->} (4,4) ; (2,1) \texttt{<-->} (2,0) ;\\
(2,1) \texttt{<-->} (2,2) ; \texttt{<ADJLIST\_END>} \texttt{<ORIGIN\_START>} (1,3) \texttt{<ORIGIN\_END>}\\
\texttt{<TARGET\_START>} (2,3) \texttt{<TARGET\_END>} \texttt{<PATH\_START>}
\end{quote}

\paragraph{Task Instructions}
You will be asked multiple questions about this maze. Please answer all questions in order. Solving one question may help with another, and you may reuse intermediate reasoning and calculations.

\paragraph{Questions}
\begin{enumerate}
    \item Is position $(0,0)$ closer to position $(2,2)$ than it is to position $(4,0)$?  
    Consider the \textbf{actual path distance through the maze}, not straight-line distance.

    \item Is position $(4,4)$ closer to position $(2,2)$ than it is to position $(0,4)$?  
    Consider the \textbf{actual path distance through the maze}, not straight-line distance.

    \item Is position $(4,0)$ closer to position $(0,4)$ than it is to position $(2,2)$?  
    Consider the \textbf{actual path distance through the maze}, not straight-line distance.
\end{enumerate}

\paragraph{Answer Format}
\begin{quote}
\ttfamily
Question 1 Answer: True/False: [reasoning] \\
Question 2 Answer: True/False: [reasoning] \\
Question 3 Answer: True/False: [reasoning]
\end{quote}

\newpage
\subsection{Sequential Questions with Point Reuse Model Performance}
\begin{table}[h!]
\centering
\caption{Results for sequential questions with point reuse}
\label{tab:experiment-2-results}
\begin{tabular}{ll|ccccc}
\toprule
\textbf{Model} & \textbf{Question} & \textbf{5×5} & \textbf{6×6} & \textbf{7×7} & \textbf{8×8} & \textbf{9×9} \\
\midrule
Gemini-2.5-Flash & Q0 & 0.94 & 0.88 & 0.78 & 0.76 & 0.70 \\
 & Q1 & 1.00 & 0.90 & 0.80 & 0.78 & 0.62 \\
 & Q2 & 0.96 & 0.92 & 0.90 & 0.78 & 0.68 \\
 & Q3 & 0.98 & 0.98 & 0.92 & 0.84 & 0.74 \\
\midrule
GPT-5-mini & Q0 & 1.00 & 0.94 & 0.92 & 0.82 & 0.70 \\
 & Q1 & 1.00 & 0.98 & 0.88 & 0.68 & 0.54 \\
 & Q2 & 1.00 & 1.00 & 0.94 & 0.68 & 0.44 \\
 & Q3 & 1.00 & 0.94 & 0.90 & 0.60 & 0.42 \\
\midrule
Claude-Haiku-4.5 & Q0 & 0.50 & 0.60 & 0.32 & 0.52 & 0.50 \\
 & Q1 & 0.46 & 0.52 & 0.44 & 0.52 & 0.58 \\
 & Q2 & 0.46 & 0.60 & 0.54 & 0.42 & 0.44 \\
 & Q3 & 0.50 & 0.60 & 0.32 & 0.52 & 0.50 \\
\midrule
DeepSeek-Chat & Q0 & 0.50 & 0.52 & 0.68 & 0.50 & 0.48 \\
 & Q1 & 0.52 & 0.54 & 0.60 & 0.60 & 0.52 \\
 & Q2 & 0.56 & 0.44 & 0.48 & 0.58 & 0.48 \\
 & Q3 & 0.84 & 0.84 & 0.74 & 0.80 & 0.70 \\
\bottomrule
\end{tabular}
\end{table}

\newpage
\subsection{Compositional Distance Comparison Model Performance}
\begin{table}[h!]
\centering
\caption{Results for compositional distance comparison accuracy by maze size and question index}
\label{tab:experiment-3-results}
\begin{tabular}{ll|ccccc}
\toprule
\textbf{Model} & \textbf{Question} & \textbf{5×5} & \textbf{6×6} & \textbf{7×7} & \textbf{8×8} & \textbf{9×9} \\
\midrule
Gemini-2.5-Flash & Q0 & 1.00 & 0.92 & 0.82 & 0.68 & 0.52 \\
 & Q1 & 0.98 & 0.88 & 0.84 & 0.68 & 0.50 \\
 & Q2 & 0.94 & 0.94 & 0.74 & 0.68 & 0.74 \\
\midrule
GPT-5-mini & Q0 & 1.00 & 0.96 & 0.44 & 0.16 & 0.04 \\
 & Q1 & 1.00 & 0.94 & 0.46 & 0.18 & 0.04 \\
 & Q2 & 0.96 & 0.92 & 0.42 & 0.16 & 0.04 \\
\midrule
Claude-Haiku-4.5 & Q0 & 0.82 & 0.66 & 0.60 & 0.58 & 0.60 \\
 & Q1 & 0.84 & 0.64 & 0.70 & 0.54 & 0.62 \\
 & Q2 & 0.76 & 0.74 & 0.52 & 0.66 & 0.70 \\
\midrule
DeepSeek-Chat & Q0 & 0.80 & 0.78 & 0.78 & 0.80 & 0.72 \\
 & Q1 & 0.86 & 0.78 & 0.82 & 0.72 & 0.62 \\
 & Q2 & 0.84 & 0.76 & 0.70 & 0.66 & 0.76 \\
\bottomrule
\end{tabular}
\end{table}

\newpage
\subsection{LLM-as-Judge Reasoning Evaluation Prompt}\label{sec:llm-as-judge-prompt}

Prompt for LLM-as-a-judge for sequential questions with point reuse task: 

\begin{quote}
\textit{Please evaluate the semantic coverage by answering the following:}

\begin{enumerate}
    \item \textbf{Information Reuse Score (0--100):} How much of the reasoning, calculations, or spatial information from Q0 is present in Q3?
    \begin{itemize}
        \item 100 = Q3 contains all key information from Q0
        \item 50 = Q3 contains about half of the information from Q0
        \item 0 = Q3 contains none of the information from Q0
    \end{itemize}

    \item \textbf{Reused Elements:} List specific elements that appear in both responses (e.g., distance calculations, path descriptions, specific coordinates).

    \item \textbf{Unique to Q0:} What information appears in Q0 but not in Q3?

    \item \textbf{Unique to Q3:} What information appears in Q3 but not in Q0?

    \item \textbf{Overall Assessment:} Brief summary of the semantic coverage and whether the model appears to be reusing information from Q0.
\end{enumerate}
\end{quote}

Prompt for LLM-as-a-judge for compositional distance comparison task: 
\begin{quote}
\textit{Please evaluate the compositional semantic coverage by answering the following:}

\begin{enumerate}
    \item \textbf{Information Reuse Score from Q0 (0--100):} How much reasoning or calculations from Q0 appear in Q2?

    \item \textbf{Information Reuse Score from Q1 (0--100):} How much reasoning or calculations from Q1 appear in Q2?

    \item \textbf{Compositional Reasoning Score (0--100):} Does Q2 show evidence of combining or composing information from Q0 and Q1?
    \begin{itemize}
        \item 100 = Clear evidence of composing both Q0 and Q1 results
        \item 50 = Some partial reuse but not clearly compositional
        \item 0 = No evidence of reusing or composing earlier results
    \end{itemize}

    \item \textbf{Reused Elements:} List specific elements that appear in Q2 from Q0 or Q1 (e.g., distance values, path descriptions, coordinates).

    \item \textbf{Unique to Q2:} What new information or calculations appear only in Q2?

    \item \textbf{Overall Assessment:} Brief summary of whether the model demonstrates compositional reasoning.
\end{enumerate}

\textit{Please provide your response in the following JSON format:}
\end{quote}

\subsection{Additional Results for Compositional Distance Comparison Task}

\begin{table}[h!]
\centering
\small
\caption{Analysis of composition distance experiment results. Q$_0$ tests the distance from corner A (top-left) to center M vs corner B. Q$_1$ tests the distance from corner D (bottom-right) to center M vs corner C. Q$_2$ tests the distance from corner B to corner C vs center M, which is designed to benefit from composing Q$_0$ and Q$_1$. $\Delta$ indicates Q$_2$ performance relative to avg(Q$_0$, Q$_1$).}
\label{tab:experiment-3-analysis-rsults}
\begin{tabular}{@{}lcccc|c@{}}
\toprule
\textbf{Model} & \textbf{Q$_0$} & \textbf{Q$_1$} & \textbf{Q$_2$} & \textbf{Overall} & \textbf{$\Delta$} \\
\midrule
Gemini-2.5-Flash & 78.8 & 77.6 & 80.8 & 79.1 & +2.6 \\
GPT-5-mini & 52.0 & 52.4 & 50.0 & 51.5 & -2.2 \\
Claude-Haiku-4.5 & 65.2 & 66.8 & 67.6 & 66.5 & +1.6 \\
DeepSeek-Chat & 77.6 & 76.0 & 74.4 & 76.0 & -2.4 \\
\bottomrule
\end{tabular}
\end{table}

\end{document}